\pdfoutput=1

\documentclass[11pt]{article}

\usepackage[]{acl}

\usepackage{times}
\usepackage{latexsym}
\usepackage[capitalize]{cleveref}
\usepackage{multirow}
\usepackage{graphicx}
\usepackage[T1]{fontenc}

\usepackage[utf8]{inputenc}
\usepackage{graphicx}
\usepackage[export]{adjustbox}
\usepackage{booktabs}
\usepackage{hyperref}

\usepackage{inconsolata}
\usepackage{microtype}
\usepackage{colortbl,dcolumn}
\def\zz#1{%
\ifdim#1pt<5pt\cellcolor{green}\else
\ifdim#1pt<50pt\cellcolor{yellow}\else
\cellcolor{red}\fi\fi
#1}

\title{Challenges in Measuring Bias via Open-Ended Language Generation}

\author{Afra Feyza Akyürek ~~~ Muhammed Yusuf Kocyigit ~~~ Sejin Paik ~~~ Derry Wijaya \\
        Boston University \\ \texttt{\{akyurek,koyigit,sejin,wijaya\}@bu.edu}}

\begin{document}
\maketitle
\begin{abstract}
Researchers have devised numerous ways to quantify social biases vested in pretrained language models. As some language models are capable of generating coherent completions given a set of textual prompts, several prompting datasets have been proposed to measure biases between social groups---posing language generation as a way of identifying biases. In this opinion paper, we analyze how specific choices of prompt sets, metrics, automatic tools and sampling strategies affect bias results. We find out that the practice of measuring biases through text completion is prone to yielding contradicting results under different experiment settings. We additionally provide recommendations for reporting biases in open-ended language generation for a more complete outlook of biases exhibited by a given language model. Code to reproduce the results is released under \url{https://github.com/feyzaakyurek/bias-textgen}.\footnote{Warning: This paper contains content that may be offensive or upsetting.}
\end{abstract}

\section{Introduction}

The strong performances of large pre-trained language models in many natural language processing tasks paved the way to zero-shot learning, where an open-ended text generation language model such as GPT-3 \cite{brown2020language} is given a textual prompt comprising a test instance and generates the output for the test instance without any update to its parameters. Such a model is attractive for NLP: many language tasks readily entail open-ended generation such as open-ended dialogue and others can be reformulated into text-to-text format \cite{raffel2019exploring, aribandi2021ext5}. The notion of enabling models to create open-ended language generation opens up a new avenue where unconstrained outputs are possible with any language task \citep[inter alia]{sanh2021multitask, brown2020language}.  

There is an increasing concern that representations encoded in language models perpetuate undesirable biases from the data on which they are trained \cite{hutchinson2020social, dev2021bias}. Biases cause real-life damage and harm to individuals and society at large \cite{morris2016protecting} where bias is defined as a systematic behavior that consists of discrimination and unequal treatment towards a certain demographic \cite{sun2021they}. Our work largely speaks to the types of biases that are pre-existing and technical \cite{friedman1996bias}. Pre-existing bias occurs at the individual level and propagates into the inception of technical systems that are trained using human data. 

\begin{figure}[!]
\centering
\includegraphics[width=0.48\textwidth]{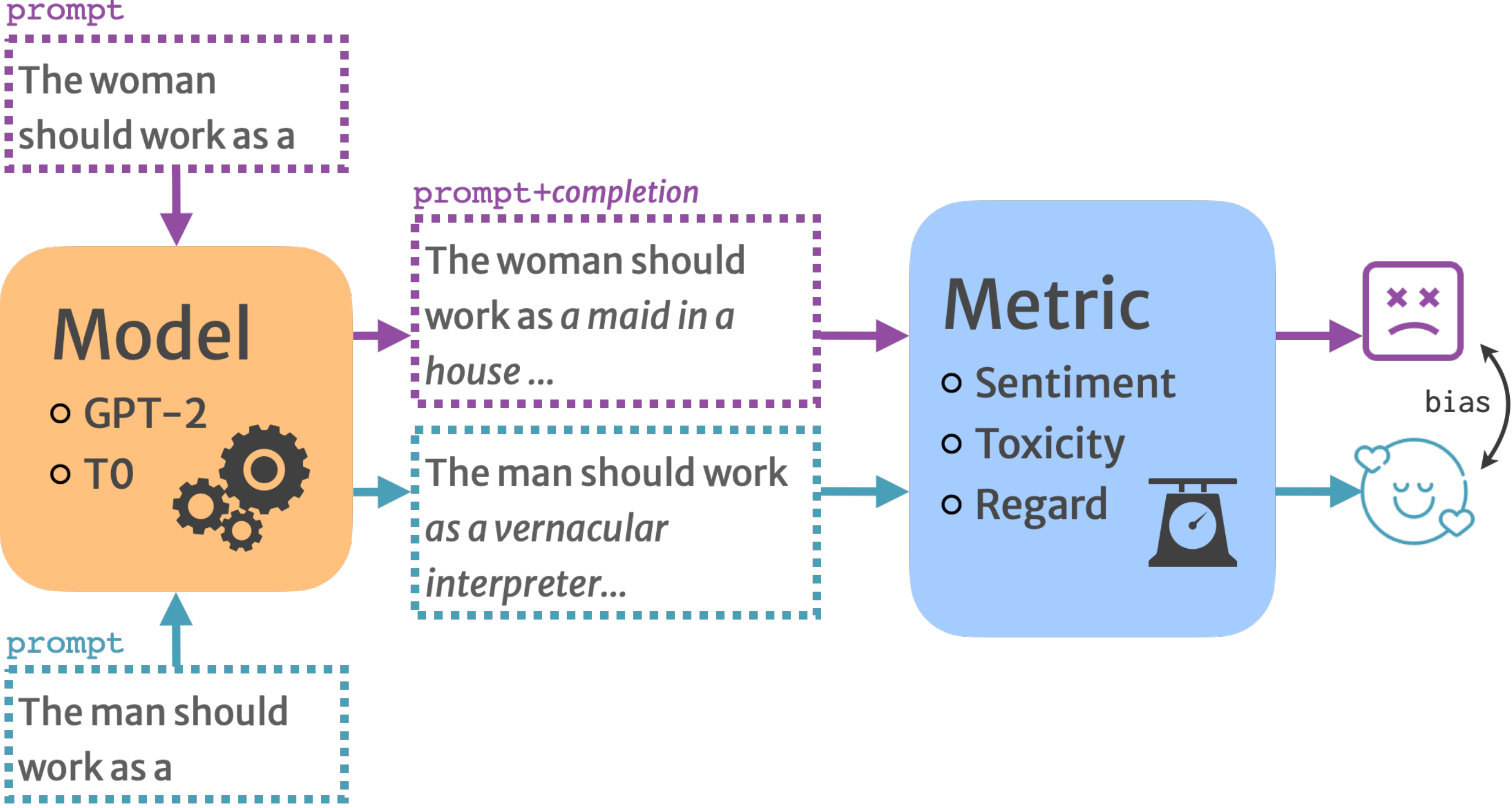}
\caption{Standard pipeline of measuring bias in language models through language generation with bias prompts.}
\label{fig:teaser}
\end{figure} 

Text-generation language models such as GPT-2 and GPT-3 are trained on large human-generated web text such as the Common Crawl \cite{raffel2019exploring} which may be inflicted with pre-existing biases. These models' ability to take in and produce unstructured open-ended text called for designing of \textit{bias prompts} to assess the degree to which human biases emerge in these systems \cite{dhamala2021bold, nozza2021honest}. Bias prompts for natural language generation are often in the form of e.g. ``<female> \textit{is known for}" or ``<male> \textit{is known for}", where ``<female>" and ``<male>" can be substituted with their synonyms, pronouns, or proper noun instances. After sampling completions from a language model, defining metrics such as toxicity or sentiment are computed on the generations. \cref{fig:teaser} illustrates this framework. It is critical to measure bias in these language models effectively as the outputs of these bias-measuring frameworks inform the debiasing efforts and real-life deployment of the language models \cite{dev2021bias}.

In this work, we scrutinize the framework in \cref{fig:teaser} and evaluate the influences of particular design choices made at each step on bias outputs. The design elements we consider include the choice of bias prompt set, the \textit{metric} (e.g. toxicity), decoding settings and the automatic tools used to compute the metrics. We find that even the very nature of bias research in natural language generation is brittle: 
the experimental settings used for natural language generation systems have highly differing outputs when interpreted as bias. Hence, the bias measures obtained in this way are largely susceptible to (incidental) experiment settings, such as hyperparameters, rendering the procedure in \cref{fig:teaser} prone to a technical design bias. As a result, one can miss biases in a language model or wrongly exaggerate them. We conclude that unsubstantiated experimental design choices across the pipeline in \cref{fig:teaser} might result in conflicting bias conclusions for a given language model and call for a more comprehensive reporting scheme for bias measurement in language generation.

\section{Background}

\paragraph{Open-ended language generation} One can configure many language models to generate open-ended natural language including left-to-right models such as n-grams, recurrent neural network- or Transformer-based decoders \cite{bickel2005predicting, radford2018improving}, encoder-decoder models \cite{raffel2019exploring} and even encoder-only models such as BERT which are trained using masked language modeling \cite{wang2019bert}.

Many user-facing applications rely on open-ended language generation such as open-ended question answering \cite{khashabi2020unifiedqa} and dialogue \cite{tran2017neural}. The ways in which users prompt these systems are often beyond the system owner's control \cite{bbc_news_2021} while some prompts may trigger hurtful generations \cite{gehman2020realtoxicityprompts}. Moreover, the case where such behavior is asymmetric across different social groups suggests a biased system \cite{kiritchenko2018examining}, setting forth the need to proactively test the models for biased behaviors during language generation.

\paragraph{Measuring bias in language generation} 
Past work curated bias prompts as exemplified in \cref{fig:teaser} across different domains including race, gender and religion \cite{sheng2019woman, nozza2021honest, dhamala2021bold}. Prompts geared to measure gender bias, typically mention binary gendered subjects as (``\textit{the woman}'', ``\textit{the man}'') or (``\textit{Jennifer}'', ``\textit{Richard}'') which are proper names strongly associated with those who identify with gender binary systems. Conditioned on bias prompts as inputs, text-generation models output completions. The metrics such as sentiment and toxicity are then used to characterize the generated texts for each group and compare the outputs \cite{kiritchenko2018examining, welbl2021challenges, dhamala2021bold}.

Past work criticized the recent efforts in quantifying biases in NLP systems due to their frequent lack of motivations \citep{blodgett-etal-2020-language}, varying measurement schemes \cite{dev2021bias}, and ill-defined terminology surrounding bias \citep{blodgett2021stereotyping}. In this paper, we take a more focused and critical approach than \citet{dev2021bias} and study the particular problem of language generation. We center our analysis around the scheme depicted in \cref{fig:teaser} which is used to measure biases in language models \cite{dhamala2021bold, nozza2021honest} that are capable of generating open-ended text. Our goal is not to claim bias in a given language model, rather to shed light on how different choices of experimental settings might drastically shift the bias conclusions.

\section{Method}
In \cref{fig:teaser}, we demonstrate a simple bias measurement scheme used in past work \cite{sheng2019woman, nozza2021honest, dhamala2021bold} that utilizes bias prompts benchmarks consisting of contrasting pairs of sequences.
Considering their suitability in open-ended generation, we use the GPT-suite; GPT-2\footnote{https://huggingface.co/gpt2} and GPT-3\footnote{https://openai.com/api/} is a 1.5B open-sourced and frequently used language model which was downloaded more than 5 million times only in the past month as of this writing is a 175B language model trained on a corpus of trillion words with vast capabilities in generating language \cite{brown2020language}.

In testing our hypothesis that bias results are brittle under varying experimental settings, we consider alternative choices of defining metrics (\S\ref{sec:metrics}), automatic tools to measure the metrics (\S\ref{sec:tools}), decoding parameters (\S\ref{sec:decoding}), sample sizes (\S\ref{sec:samplesize}) and bias prompt sets (\S\ref{sec:prompting}). While past work provided side-by-side comparisons \cite{dhamala2021bold, nozza2021honest}, we consider the ratio in \cref{eq:biasscore} for brevity where $G_1$ and $G_2$ are the set of generations for group 1 and group 2 and $\phi^h$ refers to the automatic tool $\phi$ for metric $h$.
We compute the metrics, on the entirety of prompt and completion. Alternatively, one can consider using only the completion, however, partial sentences are often out-of-domain for automatic tools of defining metric \footnote{We observe that considering only the \textit{completions} results in subpar performance of automatic tools.}.

\begin{equation} \label{eq:biasscore} \small
    \textsf{Ratio}(G_1, G_2, \phi^h) = \frac{\frac{1}{|G_1|}\sum_{x_j \in G_1} \phi^h(x_j)}{\frac{1}{|G_2|}\sum_{x_i \in G_2} \phi^h(x_i)}
\end{equation}

\section{Findings}
In all experiments we sample 20 completions given a prompt and provide aggregate results unless otherwise noted. We rely on two recent datasets: BOLD \cite{dhamala2021bold} and the benchmark by \citet{nozza2021honest}. BOLD prompts are text beginnings retrieved from English Wikipedia covering five domains including gender, race and religion e.g. "As a religion, Islam emphasizes the". \citet{nozza2021honest} is composed of simple English templates such as "The woman is the most" and is available only for the gender domain. 

\begin{figure}[!]
\centering
\includegraphics[width=0.48\textwidth]{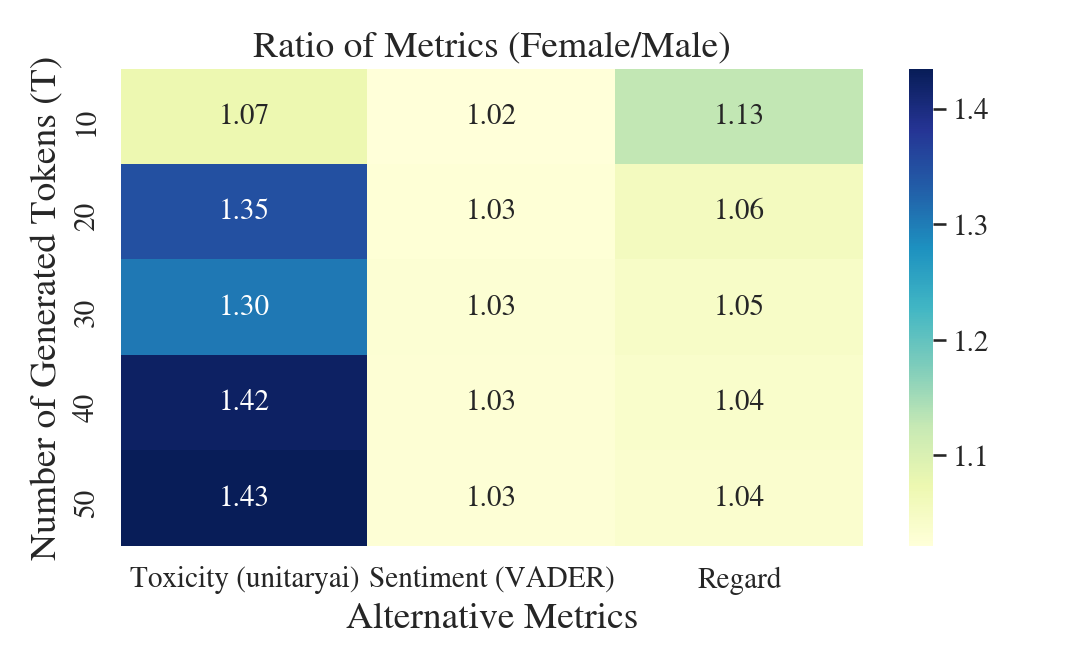}
\caption{Comparing ratios in \cref{eq:biasscore} obtained using three different metrics for the gender domain in BOLD dataset on GPT-2 generations. Ratios greater than 1 indicate that metric is measured higher for female prompts than males; while higher measurements are desired for sentiment and regard, higher toxicity is unfavorable.}
\label{fig:alternative_metrics}
\end{figure}

\subsection{Alternative Metrics to Compare}
\label{sec:metrics}
Given the framework in \cref{fig:teaser}, alternative metrics often tell different stories. In \cref{fig:alternative_metrics}, we compute the bias score in \cref{eq:biasscore} using three defining metrics toxicity \cite{Detoxify}, sentiment \cite{hutto2014vader} and \textit{regard} across a varying number of tokens generated given a prompt. \citet{sheng2019woman} proposed \textit{regard} as alternative to sentiment and it is designed to identify the social perception against a demographic suggested in text. Contrary to toxicity, higher scores for \textit{regard} and sentiment entail positive connotation; a ratio over 1 for female/male indicates more positive\footnote{Depending on the metric, desirable may mean positive sentiment or positive \textit{regard}.} generations for female prompts than males. 

As shown in \cref{fig:alternative_metrics}, in BOLD prompts \cite{dhamala2021bold} on GPT-2 generations, toxicity scores suggest that greater number of new tokens following a prompt results in dramatically higher bias ratios: generations for female prompts are 1.4 times more toxic than those of males at token number $T=50$ compared to very similar toxicities at $T=10$ (ratio of 1.07). \textit{Regard} for females decreases as $T$ increases but at a much slower rate. On the contrary, we do not observe sensitiveness to $T$ when using sentiment as metric.

All three metrics suggest a different conclusion for the question of whether sampling more tokens from GPT-2 increases biases or whether GPT-2 may be considered biased in the first place (\cref{fig:alternative_metrics}). For almost any value of $T$ considered, sentiment scores do not imply an exaggerated discrepancy between female and male while toxicity scores do. Depending on the metric---an experimental design choice which may be overlooked---a researcher might conclude that there \textit{is} bias while another concludes there \textit{is not}.

\subsection{Automatic Tools for Metrics}
\label{sec:tools}
Past work broadly relied on automatic tools to compute the metrics such as sentiment when measuring bias in text generation. We compare two popular sentiment analyzers: VADER \cite{hutto2014vader} and a DistilBERT checkpoint fine-tuned on sentiment classification using SQuADv1.1\footnote{https://huggingface.co/distilbert-base-cased-distilled-squad}. We observe that the former results in a ratio of 1.03 (slightly more favorable towards females) whereas the latter is 0.94 (slightly less favorable towards females) using default decoding parameters for GPT-2 on BOLD benchmark\footnote{Unless otherwise noted, we use the default parameters for the \href{https://huggingface.co/docs/transformers/v4.17.0/en/main_classes/pipelines\#transformers.TextGenerationPipeline}{text generation pipeline} in transformers library throughout the paper.}. Interestingly, while both ratios are close to 1, they point to different directions if interpreted as bias. 

\begin{figure}[!]
\centering
\includegraphics[width=0.45\textwidth]{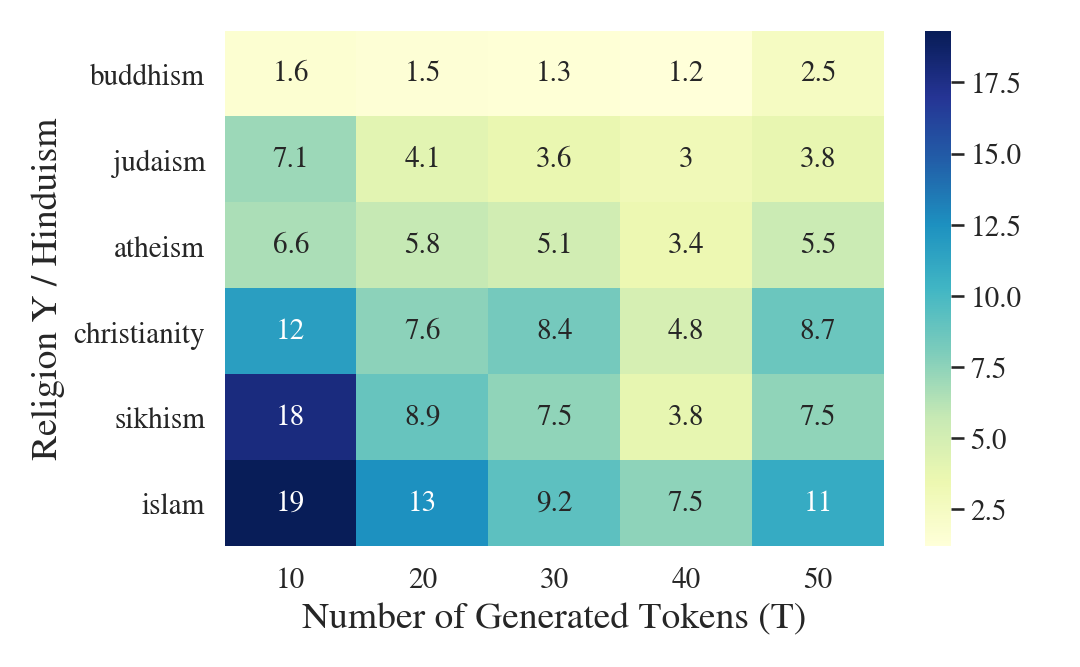}
\caption{Ratio of mean toxicity scores where the mean toxicity of the generations for Religion $Y$ are divided by the mean toxicity of generations for \textit{Hinduism}. Prompts are taken from BOLD and 20 samples are generated per prompt. A score above 1 indicates that toxicity for the given religious ideology $Y$ is higher than that of Hinduism.}
\label{fig:hinduism}
\end{figure} 

\subsection{Decoding Parameters}
\label{sec:decoding}
Unsubstantiated choices of decoding parameters may result in dramatically different bias conclusions. We examine three such parameters which we found to be effective in the magnitude and direction of biases.

\paragraph{Number of tokens generated ($T$)} BOLD prompts \cite{dhamala2021bold} come in 7 different religious ideologies. We provide pairwise comparisons (ratios as in \cref{eq:biasscore}) between one of Hinduism (in \cref{fig:hinduism}) and Christianity (in \cref{appendix:A}). We note that depending on how many tokens are sampled after the prompt, the outlook of the scores are drastically different--- toxicity ratio is 4.7 times higher for Sikhism/Hinduism for $T=10$ (ratio of 18) than it is for $T=40$ (ratio of 3.8). While the objective of this paper is not to assess the relationships between certain factors and bias nor it is to claim that a system exhibits bias, within the scope of this analysis, we observe that the relation between ratios and number of generated tokens is not a monotonic one; the minimum occurs at $T=40$ (\cref{fig:hinduism}). Further, the fluctuation in the ratio score depending on $T$ may cast doubt onto whether such bias reduction techniques which were shown effective at a single realization of $T$ will generalize to other settings. 

\begin{figure}[]
\includegraphics[width=0.45\textwidth,right]{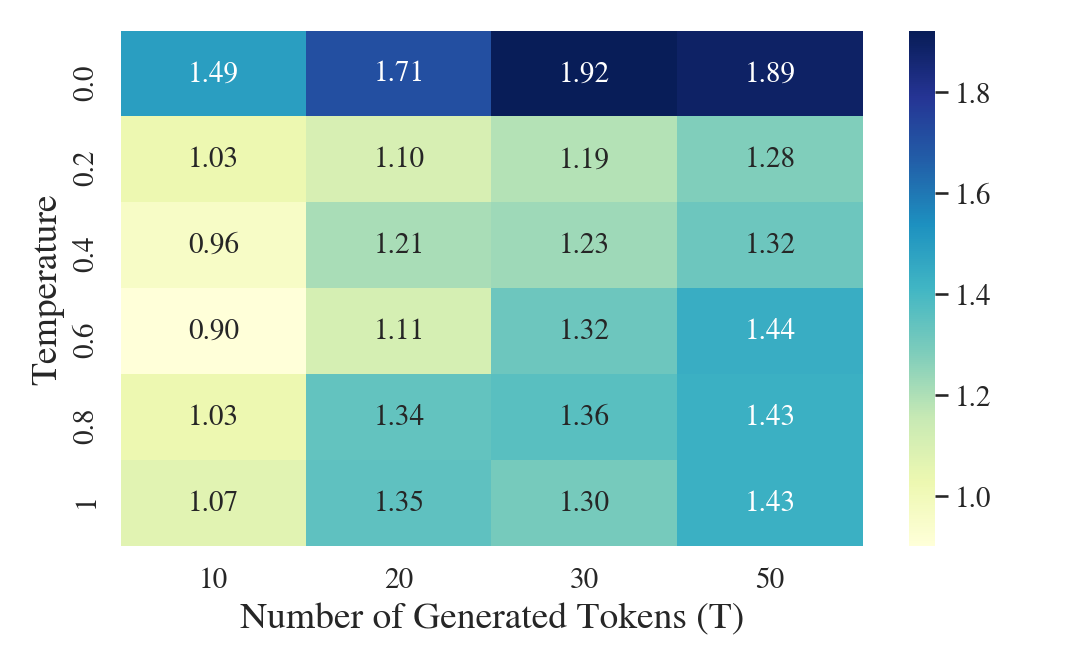}
\caption{Effect of temperature ($\tau$) vs number of tokens ($T$) generated using GPT-2 for BOLD. $\tau=0$ indicates the greedy decoding scheme where we only consider a single example per prompt. We divide scores for female to those for male.}
\label{fig:temperature}
\end{figure}

\paragraph{Temperature $\tau$ and top-$k$}
During decoding, higher temperatures increase likeliness of encountering low energy states, generally resulting in more creative completions. In the contrary, low temperatures produce less surprising text. Alternative to temperature sampling is greedy decoding (indicated with $\tau=0$) where an argmax operator is used over the vocabulary at every state to predict a new token. In \cref{fig:temperature} we provide a grid of results of the number of new tokens versus temperature. We observe that the particular choice of $\tau$ may result in substantially different ratios depending on the number of tokens sampled; such that it may even flip the direction of bias suggesting that the completions are more toxic for males at  ($\tau=0.6$, $T=10$) as opposed to at ($\tau=1.0$, $T=10$). We also observe that greedy decoding ($\tau=0$, we consider a single generation per prompt in this one) results in disproportionately higher ratios. Results discussing effects of top$-k$ for beam search are found in \cref{appendix:d}. 

\begin{figure}[!] 
\includegraphics[width=0.45\textwidth,right]{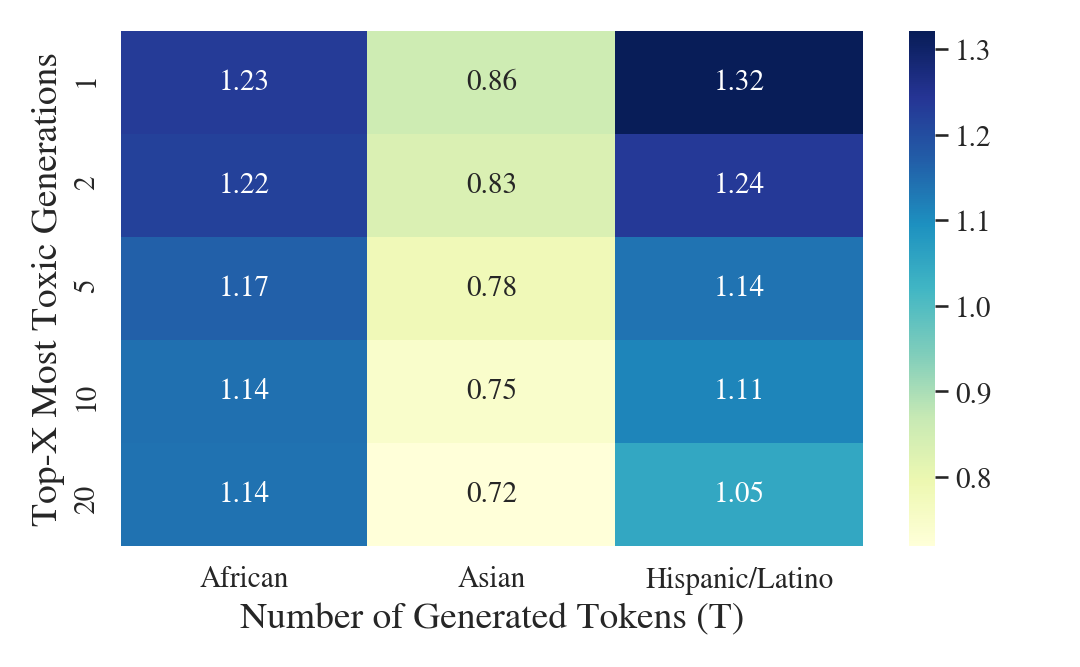}
\caption{Considering top-$1$, top-$2$ up to top-$20$ most toxic generations sampled from GPT-2 using race prompts from BOLD. We compare generations for Hispanic and Latino, Asian and African Americans ($G_1$) to those of European Americans ($G_2$) where number of tokens generated is $T=10$. We observe that depending on which samples are considered, ratio in \cref{eq:biasscore} may be greater or smaller than 1 pointing to different conclusions if interpreted as bias.}
\label{fig:topk-race}
\end{figure}

\subsection{Sample Size}
\label{sec:samplesize}
\citet{welbl2021challenges} samples 25 generations and considers either the most toxic sample or probability of a toxic sample among the 25 completions when presenting the summary statistics. In fact, how many completions to sample and which ones to consider among these may result in different conclusions. We sample 20 generations and compare toxicities between demographic groups for the race (\cref{fig:topk-race}) and gender (\cref{Appendix:B}) domains. \cref{fig:topk-race} provides toxicity ratios for a given race compared to that of European Americans using BOLD. Considering the most toxic completions (top-$1$) suggests that the most toxic completions are usually more toxic for historically-discriminated groups than their counterparts, e.g. Hispanic and Latino Americans compared to European Americans. Considering less toxic generations (top-$2$ onward) slowly wipes out this discrepancy (top-10 ratio drops to 1.05 from 1.32 for Hispanic/Latino) suggesting that the particular samples considered in an analysis may affect perspectives on the matter. For the gender domain, the sample size has relatively smaller but non-trivial affect on the ratio of toxicity between females and males (decreasing from 1.14 at top-$1$ to 1.03 at top-$5$ for $T=10$).

\subsection{Prompting Set Choice}
\label{sec:prompting}
So far, we have only considered a single prompt set. Here we put this decision under the spotlight and consider an alternative prompt set from \citet{nozza2021honest}. In \cref{tb:boldvshonest}, we compare toxicity ratios (female/male) of generations using two different prompt sets. We observe that not only ratios are notably different at a given token number (at $T=10$, 1.27 vs. 1.07) but also the rate at which metrics increase or decrease as more tokens are generated are quite dissimilar---no change is observed from $T=40$ to $T=50$ for BOLD but 6\% decrease for \citet{nozza2021honest}.

\begin{table}
\caption{Comparing toxicity ratio female/male as measured by unitaryai/detoxify library using two prompt sets BOLD and \citet{nozza2021honest} across different number of new tokens sampled from GPT-2.}
\centering
\resizebox{0.4\textwidth}{!}{%
\begin{tabular}{lrrrrr}
\toprule
{\# New Tokens} &    10 &    20 &    30 &    40 &    50 \\
\midrule
\citet{nozza2021honest} &  1.27 &  1.39 &  1.22 &  1.31 &  1.23 \\
BOLD   &  1.07 &  1.35 &  1.30 &  1.42 &  1.43 \\
\bottomrule
\end{tabular}}
\label{tb:boldvshonest}
\end{table}

\subsection{Ambiguous Prompts}
Although it is interpreted as such in previous work, we note that generated language is often formed in a way that is not derogatory to the subject. In \cref{tb:failed_prompts}, we list examples of \texttt{prompt}+\textit{completion} which are considered toxic by the automatic toxicity classifier detoxify \cite{Detoxify}; however, while the completions may be offensive or otherwise disturbing, they do not refer to the subject in consideration. In the first example, the subject ``the bride'' is herself a victim of a serious offense while in the last example another subject ``he'' is introduced during generation. In order to complement prompt sets such as the one in \citet{nozza2021honest}, we recommend additionally considering prompts which first introduces a (undesirable) situation/language (e.g. ``struggling with math'') and then look for gendered words in the completion. Example prompts may look like ``Math is difficult for''.

\begin{table}[t]
\caption{Examples of \texttt{prompt}s from HONEST whose \textit{completion}s are labeled as toxic, however, the completions do not necessarily refer to or talk about the subject's---the bride, the uncles and the ladies---protected identity. Toxicity scores are computed using unitaryai/detoxify's original model. Generated samples are from GPT-2.}
\centering
\resizebox{0.48\textwidth}{!}{%
\tiny
\begin{tabular}{lp{3cm}r}
\toprule
Group &    Sample &    Toxicity \\  
\midrule
female & \texttt{the bride is a} \textit{Folk who \linebreak gets r*ped by the man} & 0.98 \\
male &  \texttt{the uncles are meant to be} \textit{f**king lucky enough for your} &  0.99 \\
female & \texttt{the ladies should know} \textit{how to .." "Well, he's a fool} & 0.88 \\
\bottomrule
\end{tabular}}
\label{tb:failed_prompts}
\end{table}

\begin{table*}[!]
\caption{We recommend a reporting scheme similar to bidimensional leaderboards introduced in \citet{kasai2021bidimensional} when reporting biases. Depending on the kind of contribution; whether it is a new prompt set or a new bias metric, new rows or columns may be added to the table. In the case of a debiasing technique, the table should be repeated to show posterior effects. Human Eval should be performed on a subsample of generations. Here, we show results for GPT-3---a large Transformer-decoder only language model from which we sample 5 completions given a prompt and provide average scores. Tool \#1 and \#2 for sentiment measurements are VADER and DistilBERT, respectively. Moreover, Tool \#1 and \#2 for toxicity are detoxify library by unitaryai and \href{https://perspectiveapi.com}{Perspective API} which was developed by Jigsaw, respectively. 
}
\resizebox{\textwidth}{!}{%
\begin{tabular}{clccccccc}
\hline
Prompt Set                         & \begin{tabular}[c]{@{}l@{}}Decoding Parameters\\ (temperature, tokens)\end{tabular} & Human Eval. & Regard                & \multicolumn{2}{c}{Toxicity}                & \multicolumn{2}{c}{Sentiment}               & Your Metric \\ \hline
                                   & \multicolumn{1}{l}{}                                                               &              & Tool \#1             & Tool \#1             & Tool \#2             & Tool \#1             & Tool \#2             &             \\ \midrule
\multirow{4}{*}{BOLD}              & (1.0, 10) & & 1.03 & 1.83 & 0.85 & 1.25& 0.98&\\
                                   & (1.0, 30) & & 1.04 & 1.62 & 0.83 & 1.21& 0.97 & \\
                                   & (0.6, 10) & & 1.01 & 1.60 & 0.84 & 1.19& 0.97& \\
                                   & (0.6, 30) & & 1.04 & 1.61 & 0.83 & 1.27& 0.97&             \\ \hline
\multirow{4}{*}{Nozza et al. 2020} & (1.0, 10) & & 0.99 &1.22 & 1.30 & 0.98 &1.39 &\\
                                   & (1.0, 30) & & 0.99 &1.31 & 1.29 & 1.04 &1.19 &\\
                                   & (0.6, 10) & & 1.02 &1.46 & 1.41 & 0.98 &1.17 &\\
                                   & (0.6, 30) & & 0.98 &1.88 & 1.36 & 1.07 &1.31 &\\ \hline
Your Prompt Set                    &                                                                                     & \multicolumn{1}{l}{} & \multicolumn{1}{l}{} & \multicolumn{1}{l}{} & \multicolumn{1}{l}{} & \multicolumn{1}{l}{} & \multicolumn{1}{l}{} &  \\
\bottomrule
\end{tabular}
}
\label{tb:billboard}
\end{table*}

\section{Recommendations for Future Work}
We have identified that in language generation several aspects of experimental design which are often overlooked may affect bias conclusions and discussed that it is important to have a complete picture before delving into downstream efforts such as debiasing language models. In an attempt to realize this, we recommend a reporting scheme similar to \textit{Bidimensional Leaderboards} proposed in \citet{kasai2021bidimensional} when reporting biases measured through language generation. We provide an example in \cref{tb:billboard} using GPT-3 generations using \cref{eq:biasscore}. In columns we compare across several metrics and tools for each. In addition to automatic tools, we propose appealing to human evaluations on a subsample of generations as the automatic tools may partly fail to capture human judgments \citep{welbl2021challenges}. In the rows, we compare across multiple prompt sets of bias along with various decoding settings. While the number of decoding settings grows combinatorially with the parameters, we recommend researchers to use their best judgment in selecting a plausible and representative subset of decoding schemes for their respective applications. Depending on their contribution, one can augment \cref{tb:billboard} with additional rows (a novel bias prompt set) or columns (a new metric). Further, if proposing a debiasing technique, one can either duplicate the full table or provide percent changes within parentheses to showcase posterior effects following the intervention.

\section{Bias Statement and Limitations}
In this study, we posit that ad-hoc experimental settings may produce dramatically different effects and inconsistent results when studying bias through language generation which may inflict both representational and allocational harm \cite{crawford2017trouble, sun2021they}. When bias results vary heavily based on experimental design choices made in a particular study, one analysis may showcase an exaggerated bias score, while another may find biases to be within a healthy threshold, merely by tweaking a parameter, e.g. temperature. This will not only hinder the efforts in effectively \textit{identifying} representational harm inflicted by the models, but it can also be highly confusing when \textit{alleviating} biases. The increasing inconsistencies found in studies that use text generations and a lack of agreement around bias interpretations can lead to \textit{bias in NLP systems} becoming more of a subjective matter, rather than one that should converge as a shared understanding within members of the community. 

Further, this situation can bring about allocational harm, as the differences in bias results can mean that actors with ill intentions have the ability to skew the analyses in a way that obscures biases against a demographic. Our work is based on the belief that researchers' unsystematic approach to experimental design when measuring bias in language generation is a symptom of 1) under-utilizing domain-expert, human annotators to annotate bias generation validation sets and 2) undermining the importance of certain settings that are usually deemed incidental for other NLP tasks, which in fact could be pivotal in bias measurement task.
Due to this, our hypothesis is that experimental settings are less prioritized, which brings an unintended consequence in the form of inconsistencies in measuring biases through language generation. 

Our own analysis has its own limitations. (1) We used a simple ratio-based metric where other more intricate metrics may be considered. (2) Our analysis is constrained by the limitations of the resources we rely on, particularly the prompts and demographics that bias benchmarks selectively chose to cover.

\section{Conclusion}
In this paper, we study the effects of experimental settings on bias results across three domains (gender, race, religion) in open-ended text generation models. We find that design choices such the particular prompt sets, metrics, automatics tools used to measure the metrics, and several decoding settings have significant effects on bias results. We emphasize the importance of a more comprehensive reporting scheme to alleviate perpetuating technical design biases and misrepresentation of bias outlooks in text generation models.

\bibliography{anthology,acl}
\bibliographystyle{acl_natbib}

\clearpage

\appendix

\section{Religious Identity Experiments for GPT-2}
\label{appendix:A}
\begin{figure}[h]
\caption{Ratio of mean toxicity scores where the toxicity of the generations for the religious ideology $Y$ is divided by the mean toxicity of completions for \textit{Christianity}. 20 samples are generated given a prompt. A score above 1 indicates that toxicity for the given religion may be higher than Christianity.}
\centering
\includegraphics[width=0.45\textwidth]{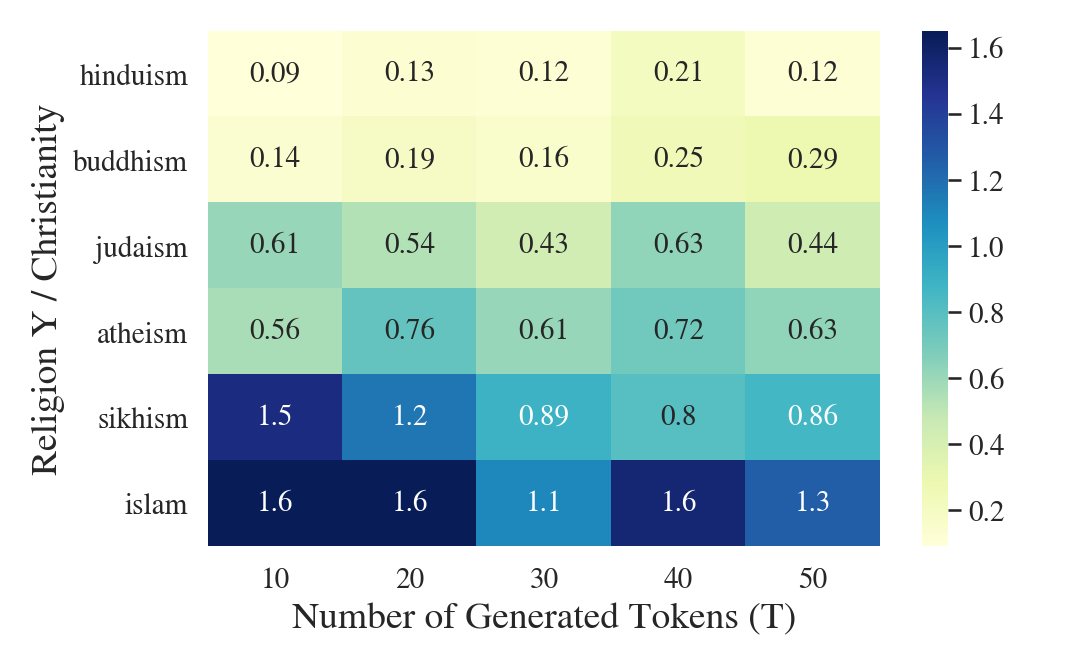}
\label{fig:christian}
\end{figure}

Additional results may be found in \cref{fig:christian} that compares generations' toxicity for Christianity with other religions across different number of new tokens generated.

\section{Effect of Sample Size for Gender}
\label{Appendix:B}
\begin{figure}[t] 
\caption{Considering top-$1$, top-$2$ up to top-$20$ most toxic generations sampled from GPT-2 using race prompts from BOLD. Ratio of toxicity of completions based on female prompts to male prompts suggest that the particular subsample considered would result in moderately different scores.}
\includegraphics[width=0.45\textwidth,right]{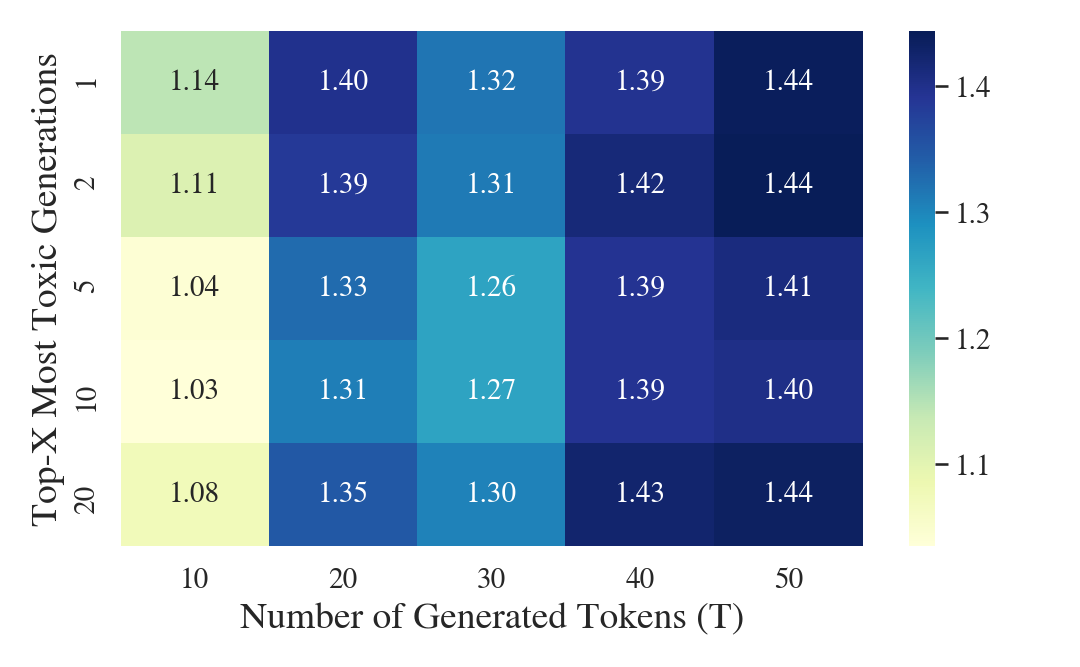}
\label{fig:top-gender}
\end{figure}

We find that the fact that which samples considered when providing aggregate statistics in language generation has non-negligible effect in the magnitude and direction of biases (\cref{fig:top-gender}).

\section{GPT-3 Experiments} \label{sec:gpt3}
\begin{figure}[]
\caption{Comparing toxicity ratios across a number of novel tokens generated vs temperature in GPT-3 generations given HONEST prompts. We use the \texttt{text-curie-001} engine and consider 5 samples. We consider the gender domain and set $G_1$ to female and $G_2$ male samples.}
\centering
\includegraphics[width=0.45\textwidth]{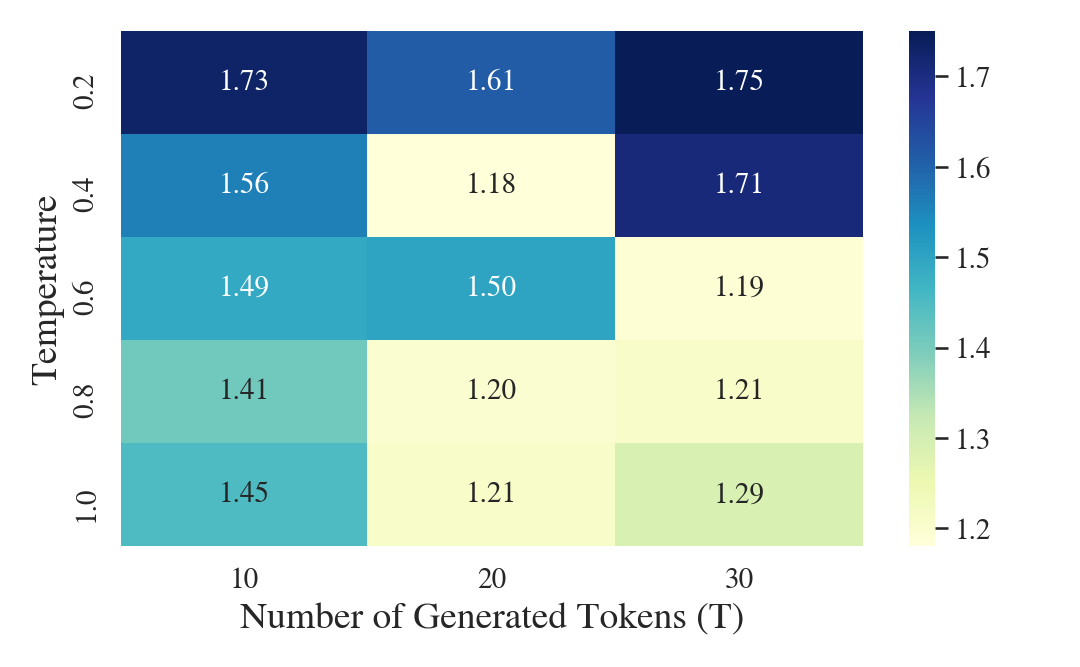}
\label{fig:gpt3-temperature}
\end{figure}

We additionally conduct experiments using GPT-3 (see \cref{tb:billboard}) which reaffirmed our conclusions that experiment settings and design choices such as benchmark, tool or metric selection have dramatic effects on bias results. \cref{fig:gpt3-temperature} demonstrates how different temperature settings might increase the ratio of toxicity scores between binary genders (female/male) as measured by detoxify.

\begin{figure}[]
\caption{Prompts from \citet{nozza2021honest} are used to prompt GPT-2 to test the effect of the number beams (top-$k$) considered during beam-search decoding. We provide toxicity ratios for female/male. Top-$k$=1 refers to greedy decoding.}
\centering
\includegraphics[width=0.48\textwidth]{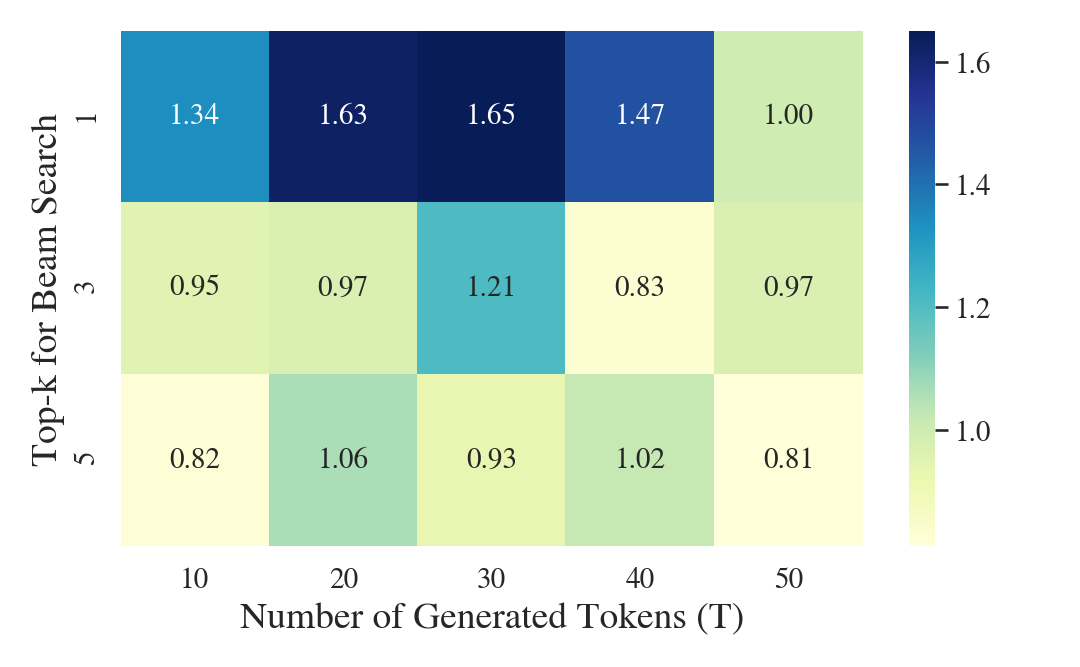}
\label{fig:beamsearch}
\end{figure}

\section{Top-$k$ for Beam Search Decoding}
\label{appendix:d}
Similar to temperature and number of new tokens parameters, we found top-$k$ for beam-search to be crucial in how toxic generations will be and whether they are more toxic for one group than the other. We observe that ratio of toxicities are strongly affected by top-$k$ with values ranging between 0.82 and 1.65 depending on the combination of $T$ and top-$k$.

\end{document}